\documentclass{article}

\usepackage{microtype}
\usepackage{graphicx}
\usepackage{subfigure}
\usepackage{booktabs} % for professional tables

\usepackage{hyperref}

\usepackage[accepted]{icml2024_arxiv}

\usepackage{amsmath}
\usepackage{amssymb}
\usepackage{mathtools}
\usepackage{amsthm}

\usepackage[capitalize,noabbrev]{cleveref}

\theoremstyle{plain}

\theoremstyle{definition}

\theoremstyle{remark}

\usepackage[textsize=tiny]{todonotes}

\usepackage{multirow} 
\usepackage{adjustbox}

\providecommand{\dataset}[0]{\mathcal{D}}

\providecommand{\sinput}[0]{\mathcal{X}}
\providecommand{\soutput}[0]{\mathcal{Y}}

\def\bx{{\boldsymbol{x}} }
\def\by{{\boldsymbol{y}} }

\DeclareMathAlphabet{\mathsfit}{\encodingdefault}{\sfdefault}{m}{sl}
\SetMathAlphabet{\mathsfit}{bold}{\encodingdefault}{\sfdefault}{bx}{n}

\newcommand{\E}{\mathbb{E}}

\newcommand{\KL}{D_{\mathrm{KL}}}

\icmltitlerunning{Uncertainty-Penalized Reinforcement Learning from Human Feedback with Diverse Reward LoRA Ensembles}

\begin{document}

\twocolumn[
% \icmltitle{Reinforcement Learning from Human Feedback with \\ Diversified Language Reward Models}

\icmltitle{Uncertainty-Penalized Reinforcement Learning from Human Feedback with Diverse Reward LoRA Ensembles}



\icmlsetsymbol{equal}{*}

\begin{icmlauthorlist}
\icmlauthor{Yuanzhao Zhai}{nudt,complex}
\icmlauthor{Han Zhang}{HIT,PCL}
\icmlauthor{Yu Lei}{PCL}
\icmlauthor{Yue Yu}{nudt,complex}
\icmlauthor{Kele Xu}{nudt,complex}
\icmlauthor{Dawei Feng}{nudt,complex}
\icmlauthor{Bo Ding}{nudt,complex}
\icmlauthor{Huaimin Wang}{nudt,complex}
\end{icmlauthorlist}

\icmlaffiliation{nudt}{National University of Defense Technology, Changsha, China}
\icmlaffiliation{complex}{State Key Laboratory of Complex \& Critical Software Environment}
\icmlaffiliation{HIT}{Harbin Institute of Technology (Shenzhen), Shenzhen, China}
\icmlaffiliation{PCL}{
Peng Cheng Laboratory, Shenzhen, China}

% \icmlcorrespondingauthor{}{}

\icmlkeywords{Machine Learning, ICML}

\vskip 0.3in
]



\printAffiliationsAndNotice{}  % leave blank if no need to mention equal contribution

\begin{abstract}
Reinforcement learning from human feedback (RLHF) emerges as a promising paradigm for aligning large language models (LLMs). 
However, a notable challenge in RLHF is overoptimization, where beyond a certain threshold, the pursuit of higher rewards leads to a decline in human preferences.
In this paper, we observe the weakness of KL regularization which is commonly employed in existing RLHF methods to address overoptimization.
To mitigate this limitation, we scrutinize the RLHF objective in the offline dataset and propose uncertainty-penalized RLHF (UP-RLHF), which incorporates uncertainty regularization during RL-finetuning. 
To enhance the uncertainty quantification abilities for reward models, we first propose a diverse low-rank adaptation (LoRA) ensemble by maximizing the nuclear norm of LoRA matrix concatenations.
% We concatenate multiple matrices of LoRA and maximize the nuclear norm to actively diversify LoRA ensembles.
Then we optimize policy models utilizing penalized rewards, determined by both rewards and uncertainties provided by the diverse reward LoRA ensembles.
Our experimental results, based on two real human preference datasets, showcase the effectiveness of diverse reward LoRA ensembles in quantifying reward uncertainty. 
Additionally, uncertainty regularization in UP-RLHF proves to be pivotal in mitigating overoptimization, thereby contributing to the overall performance.
% Experimental results on two real human preference datasets demonstrate the effectiveness of diverse reward LoRA ensembles in quantifying reward uncertainty and the efficacy of uncertainty regularization in mitigating overoptimization.
\end{abstract}

\section{Introduction}
Large language models (LLMs) possess extraordinary capacities, especially in creative content generation~\cite{2020GPT3}.
Fueled by vast corpora of internet data, which may contain low-quality and potentially biased data, LLMs can produce fabricated facts, biased or toxic text, and even content harmful to humans~\cite{perez2022red,kreps2022all}.
% Reinforcement learning from human feedback (RLHF) has emerged as a powerful technique to align LLMs with human values~\cite{2019TLDR, InstructGPT, Llama2}.
In the pursuit of addressing these issues, reinforcement learning from human feedback (RLHF)~\cite{2019TLDR, InstructGPT, Llama2} has emerged as a dominant approach in the realm of AI alignment for LLMs.

\begin{figure}[t]
    % \vskip 0.2in
    \begin{center}
    \centerline{\includegraphics[width=\columnwidth]{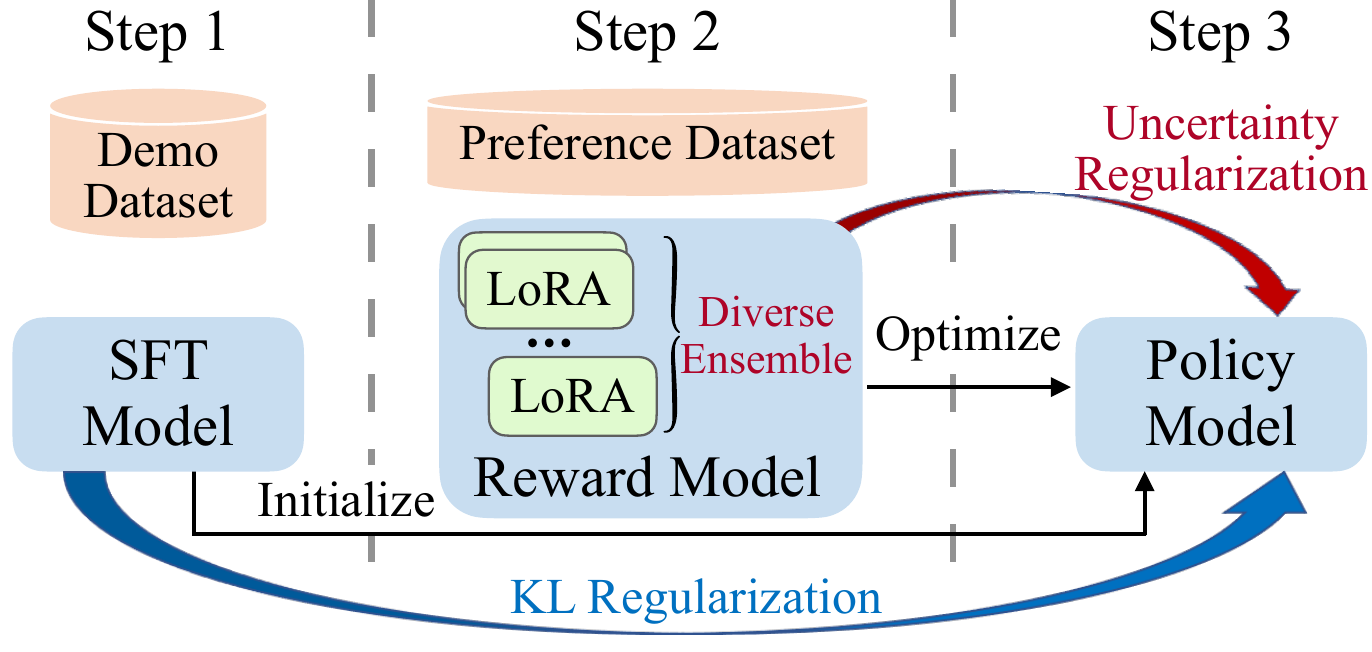}}
    \caption{Illustration of UP-RLHF. Compared to RLHF, we train diverse reward LoRA ensemble in Step 2, and add uncertainty regularization in Step 3.}
    \label{fig:Illustration}
    \end{center}
    \vskip -0.3in
\end{figure}

RLHF involves a three-step fine-tuning, as shown in Figure~\ref{fig:Illustration}. 
Step 1 contains the supervised fine-tuning (SFT) on the demonstration dataset, and reward models are trained to approximate human preferences regarding the generated output text in Step 2.
During Step 3, LLMs are conceptualized as policy models optimized by reinforcement learning (RL) algorithms, such as REINFORCE~\cite{REINFORCE}, A2C~\cite{A2C} and PPO~\cite{PPO}. 
Given prompts, LLMs are optimized to output answers that maximize scores provided by the reward model (RM).
% While successful, recent studies have brought to light potential issues that warrant further investigation.

While successful, one of the most challenging issues in RLHF is RM overoptimization~\cite{gao2023scaling}.
Overoptimization means optimizing LLMs by maximizing rewards of RM beyond a certain threshold may result in diminished human preferences, which can be approximated by the gold reward model in practice. 
Instances include generating hallucinating information to pretend expertise, or even generating overly wordy responses that can cause repeated failures~\cite{beeching2023stackllama}.
We argue that the issue is mainly caused by the overconfident RM, which is trained on limited datasets and is only an imperfect proxy for human preferences.
If an RM wrongly assigns high rewards for some out-of-distribution (OOD) samples, LLMs can be misled into outputting low-quality content.

Recent RLHF works have demonstrated the importance of introducing Kullback–Leibler (KL) penalties as regularization for mitigating the overoptimization issue~\cite{InstructGPT, Llama2, Baichuan2}.
% the regularization by Kullback–Leibler (KL) divergence, which penalizes rewards with KL penalties, can eliminate the overoptimization issue~\cite{InstructGPT, Llama2, Baichuan2}.
The intuition is adding KL regularization can regulate the output deviation of policy models from the SFT model.
% add KL penalties to rewards~\cite{InstructGPT, Llama2, Baichuan2}, regulating the deviation from the SFT model.
However, KL regularization is susceptible to overfitting~\cite{IPO}, causing a reduction in gold performance~\cite{gao2023scaling}.
% The authors in~\cite{gao2023scaling} advocate that enlarging the parameter and data size of the reward model diminishes overoptimization.
Other approaches to mitigate overoptimization include enlarging the parameter or training data size of RM~\cite{gao2023scaling}, composite RM in terms of different aspects~\cite{moskovitz2023confronting}.
% , and reward ensembles~\cite{coste2023reward, eisenstein2023helping}.
We argue that these approaches may not always be feasible because of the significantly expensive cost.

In this paper, we revisit the optimization objective of RLHF with offline datasets and show that KL regularization stemming from Step 1's demonstration dataset leads to weak regularization for low-quality OOD samples.
% We propose uncertainty regularization utilizing the reward model.
Based on this observation, we propose uncertainty-penalized RLHF (UP-RLHF), which supports additional uncertainty regularization.
We first propose the diverse reward LoRA ensemble via nuclear norm maximization in step 2.
Specifically, we concatenate multiple matrices of LoRA and maximize the nuclear norm to actively diversify LoRA ensembles.
In this manner, we train diverse LoRA ensembles, enabling reward models to have a good capability of uncertainty quantification in a parameter-efficient way.
% to enable reward models to have a good capability of uncertainty quantification
% to diversify LoRA ensembles.
Then we penalize rewards with estimated uncertainties and adopt both KL and uncertainty regularization to mitigate overoptimization.
% UP-RLHF can 
UP-RLHF can prevent LLMs from outputting high-uncertainty low-quality contents, where the KL regularization is weak, thereby mitigating the overoptimization issue.

In summary, our contributions are: 
(1) We propose UP-RLHF, which augments RLHF with uncertainty regularization by penalizing rewards with uncertainties provided by the reward model.
(2) We propose to train reward models with the diverse LoRA ensemble.
This parameter-effective approach demonstrates its effectiveness in training uncertainty-aware reward models.
% , which is cheap and introduces a few parameters.
% We propose the diverse LoRA ensemble, a novel method to train reward models with satisfactory uncertainty quantification with a few parameters introduced.
(3) Experimental results show the effectiveness of UP-RLHF in eliminating overoptimization and improving performances in terms of gold reward. 
% Under identical configurations, leveraging UP-RLHF and reward modeling with the OPT-350M matches the performance with RLHF using the OPT-2.7B reward model.
% (1) We propose to train reward models with LoRA-ensembles for uncertainty quantification with a few additional parameters.
% (2) We propose Uncertainty-penalized (UP) RLHF, a novel approach that penalizes rewards provided by the reward model with estimated uncertainties.
% (3) Extensive experimental results show the effectiveness of UP-RLHF in eliminating overoptimization and improving training stabilities.

\section{Preliminaries}
\subsection{Reinforcement Learning from Human feedback}
For an NLP task, we are given a supervised dataset $\dataset=\{(\bx^{(i)}, \by^{(i)})\}_{i=1,2,\cdots}$ of $N$ examples, 
where $\bx \in \sinput$ are prompts and $\by \in \soutput$ are the target answers.
% \subsection{RLHF Pipeline}
We outline the RLHF pipeline, which is adopted in subsequent works ~\cite{2019TLDR, InstructGPT, bai2022training}.

\textbf{Step 1: Supervised Fine-Tuning:} The initial stage commences with a pre-trained LLM, subject to fine-tuning through supervised learning, typically utilizing cross-entropy loss, with $(\bx,\by)$ samples. The outcome of this phase is denoted as $\pi^{\text{SFT}}$.

\textbf{Step 2: Reward Modeling.}  In the subsequent phase, the preference dataset with the form of $(\bx, \by^w, \by^l)$ is used to train reward models, where $\by^w$ is the one favored by the labeler and $\by^l$ is the less favored one.
Following the principles of Bradley-Terry model~\cite{bradley1952rank}, the rank loss of training the reward model is:
    \begin{equation}\mathcal{L}^{RM} = \sum_{\bx}  \log \sigma\big(r(\by^w | \bx)-r(\by^l | \bx)\big),
    \label{equation:rm}\end{equation}
where $\sigma$ is the sigmoid function. Reward model $r$ is initialized with $\pi_{\text{SFT}}$ by replacing language heads with value heads.

% the SFT model is provided with prompts $\vx$ to generate pairs of answers $\bx_{T},\boldsymbol{s'}_{T}\sim \pi_{\text{SFT}}(\cdot|\vx)$. The pairs are then subjected to evaluation by human labelers who express preferences for one answer over the other, denoted as $\bx_{T} \succ \boldsymbol{s'}_{T} |\vx$, where $\bx_{T}$ is favored, and $\boldsymbol{s'}_{T}$ is the less favored one.
% Following the principles of the Bradley-Terry model~\cite{bradley1952rank}, the human preference distribution $p^*$ can be expressed as:
%     \begin{align*}
%         p^*(&\vy_1\succ\vy_2|\vx) = \frac{1}{1+\exp\left(r^*(\vx,\vy_1)-r^*(\vx,\vy_2)\right)}. 
%     \end{align*}
    % Assuming the access to a dataset $\{(\vx^{(i)},\vy^{(i)}_w,\vy^{(i)}_l)\}_{i=1}^N$ sampled from $p^*$. We parameterize the reward as $r$ and estimate it via maximum log-likelihood:

    % Constraints like $\E\left[r(\vy|\vx)\right] = 0$ might be incorporated to lower the variance.
    
\textbf{Step 3: RL Fine-Tuning.} 
% We define generation tasks as a Markov Decision Process (MDP)  $\langle \xSpace, \ySpace,\rewardfunc, \transFnDef, \hor\rangle$ using a finite vocabulary $\vocab$.
For a prompt $\bx$ sampled from the dataset $\dataset$, the language model to be optimized is denoted as $\pi_\theta$, which generates the target answer $\by$.
The transition function deterministically appends an answer $\by$ to the end of the prompt $\bx$.
Then the learned reward model provides a trajectory-wise reward $r(\by|\bx)$.
Prior works formulate the optimization problem as:
% \begin{align}
% \max_{\pi_\theta} & \E_{\vx\sim \mathcal{D}, \vy\sim \pi_\theta(\cdot|\vx)}\left[r(\vy|\vx) - \\ & \beta\KL(\pi_\theta(\cdot|\vx)\Vert \pi_{\text{SFT}}(\cdot|\vx))\right]
% \label{equation:rl_reward}
% \end{align}
\begin{align}
\mathop{\mathrm{arg \ max}}_{\pi_\theta} \  & \E_{\bx \sim \dataset, \by \sim \pi_\theta(\cdot|\bx)}\left[ r(\by|\bx) - \right. \nonumber \\
& \left. \beta \log(\pi_\theta(\by|\bx)/ \pi^{\text{SFT}}(\by|\bx))\right],
\label{equation:rlhf_reward}
\end{align}
% where $d_\pi_\theta(\bx)$ is the state occupancy measure which depends on the initial states $\bx_0 \sim \dataset$ and policy $\pi_\theta$, $\beta$ is a constant, and $\bI_{T}$ denotes the indicator functions of the state sets at the time step $T$.
where $\beta$ controls the strength of the KL penalty.
The KL penalty $\beta \log(\pi_\theta(\by|\bx)/ \pi^{\text{SFT}}(\by|\bx))$ is used to regulate the deviation from the SFT model.
% , which is important for training stability and reducing reward hacking~\cite{Llama2}. 
% In practice, the KL penalty is typically estimated by penalizing rewards of each token with $\beta\log\left(\pi_\theta(a|\bx)/ \pi_{\text{SFT}}(a|\bx)\right)$.
    % The standard approach is to directly employ PPO \citep{schulman2017proximal, ouyang2022training} to optimize the modified reward $r(\vy|\vx)-\beta\left(\log\pi_\theta(\vy|\vx)-\log \pi_{\text{SFT}}(\vy|\vx)\right)$.
Existing works utilize RL algorithms~\cite{InstructGPT, Llama2, ReMax}, typically PPO~\cite{PPO}, to solve objective~\ref{equation:rlhf_reward}.
% Besides, deriving from objective~\ref{equation:rlhf_reward}, DPO~\cite{DPO} leverages rank loss to bypass the need for RL and explicit reward modeling.

\subsection{Low-Rank Adaptations}
% LLMs are usually composed of $L$ stacked transformer blocks~\cite{vaswani2017attention}. Each block comprises two sub-modules: a multi-head attention (MHA) and a fully connected network. For an input sequence $X \in \mathbb{R}^{n \times d}$, $h$ attention heads are computed in parallel by the MHA.
As one of the most popular Parameter-Efficient Fine-Tuning (PEFT) methods, LoRA~\cite{LoRA} introduces bypass modules to update pre-trained models through up-down projection, involving down-projection matrices denoted as $A$ and up-projection matrices denoted as $B$.
Throughout fine-tuning, the model initiates with fixed pre-trained weights $W^{(0)}$ and evolves to $W = W^{(0)} + \Delta W$.
For each LoRA unit, the forward pass can be expressed as:
\begin{align}
z^{out}=W^{(0)}z^{in}+\Delta W z^{in}=W^{(0)}z^{in}+BAz^{in},
\end{align}
where $z^{in}, z^{out} \in \mathbb{R}^{n \times d}$ are inputs and outputs of transformer layers, $W, W^{(0)}, \Delta W\in \mathbb{R}^{d \times d}, A \in \mathbb{R}^{r \times d}$ and $B \in \mathbb{R}^{d \times r}$ with $r \ll d$. 
During the initiation of training, random Gaussian initialization is applied to $A$, while $B$ is initialized to zero.
LoRA introduces significantly fewer trainable parameters, often less than 1\% of the original model size.

\section{Methods}

\subsection{Analysis of Regularizations in RLHF}
RLHF can be formulated as reverse RL with offline datasets $\dataset$.
We theoretically analyze its overall objective which is intractable, and show how to optimize it approximately.
% how the KL penalty term is derived and what approximations it makes.
Recall our original goal is to find a policy that maximizes the expected trajectory-wise reward:
\begin{equation}
\mathop{\mathrm{arg \ max}}_{\pi_\theta} \E_{(\bx, \by) \sim \rho_{\pi_\theta}} r(\by|\bx),
\label{equ:original_objective}
\end{equation}
where $\rho_{\pi_\theta}$ is the occupancy measure which depends on the policy ${\pi_\theta}$.
Optimizing Equation~\ref{equ:original_objective} poses challenges attributable to the interdependence of $\rho_{\pi_\theta}$ and $\pi_\theta$, compounded by the necessity to gather samples from $\pi_\theta$.
With the first-order approximation of the objective \cite{TRPO,AWR}, we can formulate the following constrained policy optimization problem:
\begin{equation}
\begin{aligned}
    \mathop{\mathrm{arg \ max}}_{\pi_\theta} &  \int_{\bx}  \dataset(\bx) \int_{\by}   \pi_\theta(\by|\bx)  r(\by|\bx) \ d\by\ d\bx \\
    \textrm{s.t.}  \quad  \int_{\bx} & \dataset(\bx) \ \mathrm{D_{KL}} \left( \pi_\theta(\by |\bx) || \pi_{\dataset}(\by |\bx) \right) d\bx \leq \epsilon, \label{eqn:constrain}
\end{aligned}
\end{equation}
where $\pi_{\dataset}$ is the behavior policy induced by $\dataset$.
The constraint in Equation~\ref{eqn:constrain} ensures that the new policy $\pi_\theta$ is close to the data distribution of $\pi_{\dataset}$, and therefore the surrogate objective remains a reasonable approximation. 

Forming the Lagrangian of the constrained optimization problem presented above, we obtain the loss function:
\begin{equation}
 \begin{aligned}
     \mathcal{L}_\theta  =  \int_{\bx} & \dataset(\bx) \int_\by  \pi_\theta(\by | \bx)  r(\by|\bx) \ d\by\ d\bx  \\  + \beta &\left(\int_{\bx} \dataset(\bx) \mathrm{D_{KL}} \left( \pi_\theta(\by |\bx) || \pi_{\dataset}(\by |\bx) \right) d\bx \right) ,
     %& + \int_{\bx} \alpha_\bx \left(1 - \int_\rva \ \pi_\theta(\rva | \bx) d\rva \right) d\bx ,
     \label{eqn:Lagrangian}
 \end{aligned}
 \end{equation}
where $\beta$ is a Lagrange multiplier. Upon differentiating the objective function $\mathcal{L}(\pi_, \beta)$ with respect to $\pi_\theta(\by|\bx)$ and subsequently solving for the optimal policy $\pi^\star$, the resultant expression for the optimal policy is as follows:
\begin{equation}
 \begin{aligned}
     \pi^\star(\by|\bx)  =  \frac{1}{Z(\bx)} \ \pi_{\dataset}(\by|\bx) \ \mathrm{exp}\left(\frac{1}{\beta} (r(\by|\bx) \right),
 \end{aligned}
 \label{eqn:optimal_policy}
\end{equation}
where 
\begin{equation}
 \begin{aligned}
Z(\bx) =\sum_{\by}\pi_{\dataset}(\by|\bx) \exp\left(\frac{1}{\beta}(r(\by|\bx)\right) \nonumber
 \end{aligned}
 \label{eqn:partion}
\end{equation}
is the partition function or normalizing constant.
Following~\cite{korbak2022reinforcement, go2023aligning}, we utilize the reverse KL divergence between $\pi_\theta$ and $\pi^\star$ for distribution matching:
% \begin{align*}
%         \KL(\pi_\theta, \p^\star) = \E_{s \sim \dataset} \log  \frac{\pi_\theta(x)}{\pi^\star} 
%   = \log Z -   \frac{1}{\beta}\, \E_{s \sim \dataset} \beta\left(\log\pi_\theta(\bx)/ \pi_{\text{SFT}}(\bx)\right),  \label{eq:klrmax}
% \end{align*}
% TODOOOOOOOOOOOOOO
%%%%%%感觉Token-wise到这写不下去了%%%%%%%%%
% \begin{align*}
% \KL(\pi_\theta, \pi^\star) &= \E_{s \sim \dataset} \log \frac{\pi_\theta(x)}{\pi^\star} \\
% &= \log Z - \frac{1}{\beta}\, \E_{s \sim \dataset} \beta\left(\log\frac{\pi_\theta(\bx)}{\pi_{\text{SFT}}(\bx)}\right),
% \end{align*}
\begin{equation}
\begin{aligned}
\KL(\pi_\theta, & \pi^\star) = \E_{\bx \sim \dataset}\E_{\by \sim \pi_\theta(\by|\bx)} \log \frac{\pi_\theta(\by |\bx)}{\pi^\star(\by |\bx)} 
\\   =  - &\frac{1}{\beta}\,\E_{\bx \sim \dataset}\E_{\by \sim \pi_\theta(\by|\bx)}   (r(\by|\bx) \\  &  - \beta \log\frac{\pi_\theta(\by |\bx)}{\pi_{\dataset}(\by |\bx)} - \beta\log Z(\bx) ),
\label{eqn:real_obj}
\end{aligned}
\end{equation}

Following the analysis of previous works~\cite{AWR,zhu2023fine}, the partition function $Z(\bx) \approx 1$.
According to Equation~\ref{eqn:real_obj}, minimizing $\KL(\pi_\theta, \pi^\star)$ coincides with the objective:
\begin{equation}
\begin{aligned}
\mathop{\mathrm{arg \ max}}_{\pi_\theta} \  & \E_{\bx \sim \dataset}\E_{\by \sim \pi_\theta(\by|\bx)} \left[r(\by|\bx) - \right.  \\
  & \beta \log (\pi_\theta(\by|\bx)/ \pi_{\dataset}(\by|\bx))].
\label{equation:before_approximates}
\end{aligned}
\end{equation}

% \begin{align}
% \mathop{\mathrm{arg \ max}}_{\pi_\theta} \  & \E_{\bx \sim \dataset}\E_{\by \sim \pi_\theta(\by|\bx)} \left[r(\by|\bx) - \right. \nonumber \\
%   \lambda_1 \log &(\pi_\theta(\by|\bx)/ \pi_{\dataset}(\by|\bx))+\lambda_2 u(\by|\bx)].
% \label{equation:target_reward}
% \end{align}

% \begin{align}
% \mathop{\mathrm{arg \ max}}_{\pi_\theta} \  & \E_{\bx \sim \dataset}\E_{\by \sim \pi_\theta(\by|\bx)} \left[r(\by|\bx) - \right.\lambda u(\by|\bx)].
% \label{equation:target_reward}
% \end{align}

% We note that there are two intractable terms to optimize Equation~\ref{equation:target_reward}.
We note that $\pi_{\dataset}$ is intractable to obtain, as the generation of ${\dataset}$ can be diverse, e.g., by either $\pi^{\text{SFT}}$, powerful LLMs like GPT-4, or humans. Therefore, the distribution of the behavior policy $\pi_{\dataset}$ is not accessible. 
% The other is $Z(\bx)$, as estimating it requires multiple samples.
Since $\pi^{\text{SFT}}$ has been fine-tuned on part of $\dataset$,
we can approximate $\pi_{\dataset}$ with $\pi^{\text{SFT}}$ and then obtain the objective as in Equation~\ref{equation:rlhf_reward}.

% However, only using KL regularization has drawbacks.
% Considering an answer $\by$ with a low reward, according to the optimal policy~\cite{eqn:optimal_policy}, the probability of generating $\by$ would be small.
% According to the optimal policy~\cite{eqn:optimal_policy}, if $\by$ has a low reward, the probability of generating $\by$ would be small.

Considering a low-quality answer $\by$, even if its generation probability is small for a satisfactory policy model~\ref{eqn:optimal_policy}, we may sample such $\by$ during RL training.
In this case, the KL penalty in Equation~\ref{equation:rlhf_reward} becomes weaker or even negative, which would cause overoptimization.
This problem would be exacerbated when the RM wrongly assigns high rewards for such OOD low-quality samples.
% excessively differentiates between positive and negative samples, i.e. large $r(\by^w | \bx)-r(\by^l | \bx)$.
% However, the approximated objective may cause over-optimization~\cite{gao2023scaling}.

Trained on $\dataset$, reward models should be well-calibrated and be greatly uncertain for OOD $(\bx,\by)$ samples, which correspond to small $\pi_\dataset(\by|\bx)$.
% larger KL divergence between $\pi_\theta$ and $\pi_{\dataset}$.
Given an answer $\by$ generated by $\pi_\theta(\by|\bx)$, the more OOD the sample is, the larger the penalty term should be.
Therefore, we can approximate the intractable term in \ref{equation:before_approximates} with the uncertainty estimation of reward models $u(\by|\bx)$, which induces the following objectives:
\begin{equation}
\begin{aligned}
\mathop{\mathrm{arg   \ max}}&_{\pi_\theta} \  \E_{\bx \sim \dataset}\E_{\by \sim \pi_\theta(\by|\bx)} \left[r(\by|\bx) - \right.  \\
  \beta_1 &  \log (\pi_\theta(\by|\bx)/ \pi_\text{SFT}(\by|\bx)) - \beta_2 u(\by|\bx)],
\label{equation:after_approximates}
\end{aligned}
\end{equation}
where $\beta_1$ and $\beta_2$ are coefficients to control the KL and uncertainty regularization respectively.

% We logarithmically transform both sides of Equation~\ref{eqn:optimal_policy} and derive the following expression:
% \begin{equation}
%     \log Z(\bx) = \frac{1}{\beta} r(\by|\bx) - \log \frac{\pi^\star(\by |\bx)}{\pi_{\dataset}(\by |\bx)}. 
%     \label{eqn:partion_function}
% \end{equation}

% Since directly calculating $Z(\bx)$ is intractable, we can estimate it according to Equation~\ref{eqn:partion_function}.
% From Equation~\ref{eqn:partion_function}, we observe that $\log Z(\bx)$ 

% \subsection{Relationship with PPO and DPO}

\subsection{Training Diverse Reward LoRA Ensembles}
% Deep ensemble is the most popular
To estimate the reward uncertainty $u(\by|\bx)$, we investigate the ensemble approach, which is widely adopted for enhancing the uncertainty of deep learning methods. 
% Deep ensembles, as proposed by Lakshminarayanan et al. (2017), provide a practical alternative to the fully Bayesian treatment of Bayesian neural networks (Neal, 2012) or Monte-Carlo Dropout (Gal & Ghahramani, 2016).
Since reward models (RM) are also initialized from LLMs, we train multiple LoRAs instead of reward models for ensembles, which is more parameter-effective.
Then the forward pass can be formulated as:
\begin{equation}
\begin{aligned}
z^{out} &= \frac{1}{N} \sum_{n=1}^{N} (W^{(0)}x+\Delta W_n z^{in}) \\
&=\frac{1}{N} \sum_{n=1}^{N} (W^{(0)}x+B_n A_n z^{in}),
\end{aligned}
\end{equation}
where $\Delta W_n$ are different LoRAs of the ensemble.
Though LoRA-ensemble members have random initialization, we observe that LoRA ensembles can not exhibit satisfactory uncertainty quantification abilities
% are not enough in terms of uncertainty quantification and model calibration.
We hypothesize this is due to a lack of diversity between LoRA ensembles.
Recall that LoRA only learns parameter-update, the output of different ensemble members can be more homogeneous compared to traditional deep ensembles.
Similar phenomena are also observed in other fine-tuning methods of LLMs' ensembles~\cite{gleave2022uncertainty, eisenstein2023helping}.

\begin{figure}[h]
% \vskip 0.2in
\begin{center}
\centerline{\includegraphics[width=\columnwidth]{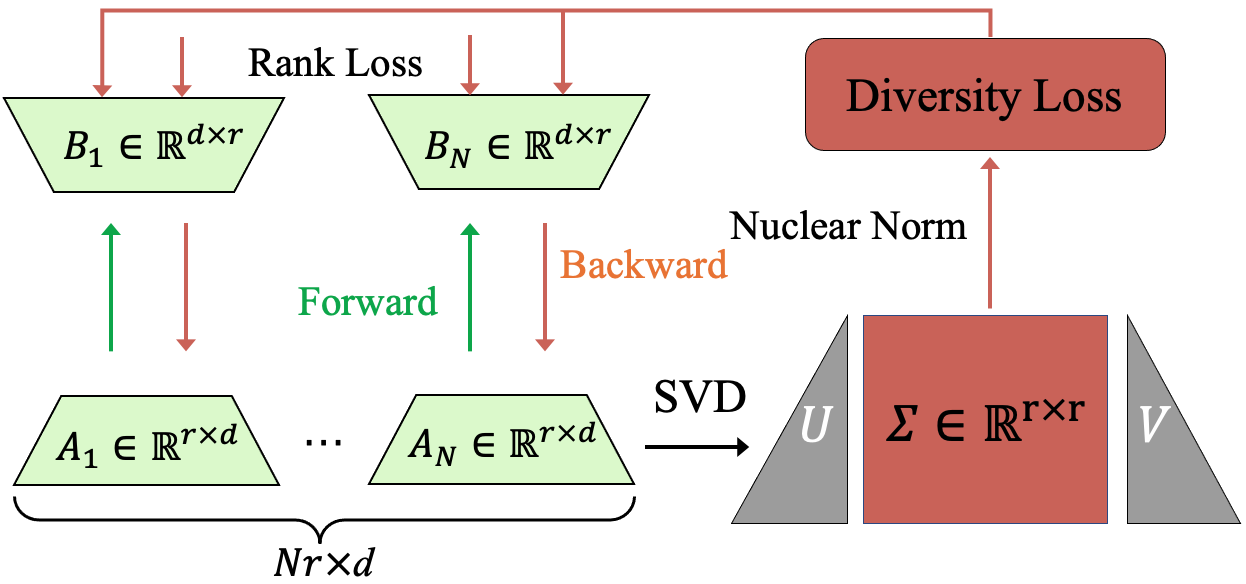}}
\caption{Illustration of training diverse reward LoRA ensembles.}
\label{fig:NNM-LoRA-Ensemble}
\end{center}
\vskip -0.2in
\end{figure}

To actively diversify reward LoRA ensembles, we propose a diversity regularization via Nuclear Norm Maximization when training LoRA ensembles.
As shown in Figure~\ref{fig:NNM-LoRA-Ensemble}, we first concatenate multiple $A_n$ along the LoRA dimension $r$ and obtain matrix $A \in \mathbb{R}^{Nr \times d}$.
If LoRA-ensemble members are totally homogeneous, the rank of $A$ equals the rank of LoRA member $A_n$. 
On the contrary, diverse members mean linearly independent along the first dimension of $A$.
Therefore, we could measure the diversity (or the homogeneity) of the LoRA ensemble with the matrix rank of the matrix $A$.
% Since the rank optimization problem is known to be NP-hard, we utilize a convex surrogate, nuclear norm to replace the rank.
Since the rank optimization problem is known to be NP-hard, we leverage the convex surrogate, nuclear norm, as a computationally efficient approximation of matrix rank, which is calculated via singular value decomposition (SVD).
In addition to the rank loss in Equation~\ref{equation:rm}, the loss function of training diverse reward LoRA Ensemble is:

% \begin{equation}\mathcal{L}_R = \sum_{\bx}  \log \sigma\left(r(\by^w | \bx)-r(\by^l | \bx)\right) + ,
% \label{equation:NNMLoRAEnsemble}\end{equation}

\begin{align}
\mathcal{L}^{RM} = &\underbrace{{\sum_{\bx}  \log \sigma  \big( \frac{1}{N} \sum_{n=1}^{N} r_n(\by^w | \bx)-  \frac{1}{N} \sum_{n=1}^{N} r_n r(\by^l | \bx) )}}_{\text{Rank loss}}  \nonumber  \\  &+ \underbrace{ \lambda \frac{1}{M} \sum_{m}^{M} \Vert A \Vert _ */{\Vert A \Vert _ F} }_{\text{Diversity regularization}},
\end{align}
where $\lambda$ is the NNM weight to control the diversity loss, $\Vert A \Vert _ *$ is the nuclear norm of $A$, and ${\Vert A \Vert _ F}$ is the Frobenius norm of $A$, which is used to control the value of weights not to be too large.

After training reward models with the diverse LoRA ensemble, we can estimate the reward uncertainty using the standard deviation:

\begin{align}
u(\by|\bx) = \sqrt{\frac{1}{N} \sum_{n=1}^{N} \big (r_n(\by|\bx) - \frac{1}{N} \sum_{n=1}^{N} r_n(\by|\bx) \big)^2}.
\end{align}

% We propose the Nuclear-Norm Maximization (NNM) LoRA ensemble for training reward models, which learns diverse ensembles while introducing a few parameters.

\subsection{Overall Optimization Objectives}
In Equation~\ref{equation:after_approximates}, three scalars including reward, KL penalty, and uncertainty penalty are to be optimized with the RL objective.
To prevent the three terms from interfering with each other, we make the KL regularization independent of the actor loss.
Specifically, we only optimize the uncertainty penalized rewards using RL algorithms:

\begin{equation}
\begin{aligned}
% \mathop{\mathrm{arg \ max}}_{\pi_\theta} \ 
\mathcal{J}^{RL}_{\theta} = \E_{\bx \sim \dataset} & \E_{\by \sim \pi_\theta(\by|\bx)}  \big[r(\by|\bx) \\ & - \beta_2 \big(u(\by|\bx) - \bar{u}(\by|\bx) \big) \big],
\label{equation:RL}
\end{aligned}
\end{equation}
where $\bar{u}(\by|\bx)$ represent the uncertainty of rewards models for $(\bx,\by)$ due to the different scales of ensemble members.
In practice, we use the mean uncertainty of all previously seen samples to approximate $\bar{u}(\by|\bx)$.

% Since the KL penalty is differentiable, t
For KL regularization, the objective is:
\begin{align}
\mathcal{J}^{KL}_{\theta}=  -\beta_1 \E_{\bx \sim \dataset}\E_{\by \sim \pi_\theta(\by|\bx)} [(\log \frac{\pi_\theta(\by|\bx)}{\pi_{\text{SFT}}(\by|\bx)})^2 ], 
\label{equation:KL}
\end{align}
where we utilize the KL estimator with lower variance, low bias, and positive assurance.
Since the objective~\ref{equation:KL} is differentiable, we directly optimize it via gradient descent. 
% Then we plus the KL loss to  $\mathcal{L}_{\theta}^{\text{RL}}$ of uncertainty penalized RL in \ref{equation:RL}, and obtain the
Overall, the objective of UP-RLHF is as:
\begin{align}
\mathcal{J}_{\theta}^{\text{UP-RLHF}} = \mathcal{J}_{\theta}^{\text{RL}} + \mathcal{J}_{\theta}^{\text{KL}}.
\end{align}
The KL regularization can be seen as the regularization from step 1 of the RLHF pipeline, while the uncertainty penalty can be seen as the regularization from step 2.

\section{Experimental Results}
In this section, we conduct empirical experiments to evaluate the alignment of UP-RLHF on two extensively utilized RLHF tasks, namely summarization and question-answering. 
% The results reveal that P3O outperforms various robust baselines in terms of both the KL-Reward trade-off and the quality of generated content.
We aim to investigate three primary research questions (RQs):
\begin{itemize}
    \item RQ1 (Step 2: Reward modeling): How well does diverse reward LoRA Ensemble improve the uncertainty quantification of reward models?
    \item RQ2 (Step 3: RL Fine-Tuning): How well does uncertainty penalization mitigate the overoptimization issue?
    \item RQ3 (Performance): How does UP-RLHF perform compared to existing RLHF methods?
\end{itemize}

To answer the above questions, we will first provide a concise introduction to the datasets and training setups.
The subsequent discussion includes evaluations of both reward models and policy models.
% , followed by the ablation studies of our methods.
% , and the evaluation metrics employed in our experiments, followed by a detailed elaboration of these findings.
\subsection{Datasets and Training Setups}
% \subsubsection{}
\textbf{Datasets.} 
% Our experiments focus on two extensively utilized RLHF tasks, namely summarization and question-answering. 
For the summarization task, we employ the ``TL;DR'' (Too Long; Didn't Read) dataset introduced by Völske et al. (2017). 
In this dataset, $\bx$ represents a forum post sourced from Reddit, and $\by$ corresponds to the respective summary. 
Notably, we use the gold reward to relabel the dataset in terms of preference, ensuring that the gold reward is the perfect proxy for the relabeled dataset.

In the question-answering task, following prior work, we use the Anthropic Helpful dataset~\cite{bai2022training} with human preference without additional relabeling.
$\bx$ signifies a fragment of a conversation involving interactions between a human and a digital assistant. The model is specifically trained to generate the helpful subsequent turn of the assistant, denoted as $\by$.

% \texttt{CarperAI/openai\_summarize\_tldr\_rm\_checkpoint}

\textbf{Training Setups.}
In the summarization task, the policy model is established using OPT-1.3B~\cite{OPT}, and the reward model is established using OPT-350m.
In the question-answering task, both the policy model and the reward model are established using Llama2-7B~\cite{Llama2}.

According to the scaling law of the reward model~\cite{gao2023scaling}, RMs with larger parameters and more training data are more robust to optimization.
Therefore, we use fine-tuned GPT-J-6B~\footnote{huggingface.co/CarperAI/openai\_summarize\_tldr\_rm\_checkpoint} as the gold reward model in the summarization task because of its larger RM parameter size and satisfactory accuracy (75\% on the test set).
In the context of the question-answering task, 3B SteamSHP-XL~\footnote{huggingface.co/stanfordnlp/SteamSHP-flan-t5-xl} is chosen as the gold reward model because of its larger RM training data size than the reward model, which is fine-tuned on both the HH and SHP~\cite{pmlr-v162-ethayarajh22a} datasets.

Following~\cite{yao2023deepspeed}, for both tasks, we perform a random partitioning for the datasets into three segments: 20\% for step 1, 40\% for step 2, and the remaining 40\% for step 3. 
% All experiments were conducted using four different random seeds, and the reported results are the averages.
% More experiment details for both Step 2 and Step 3 can be found in Appendix~\ref{appendix:exp}.

\subsection{Reward Model Evaluation}
\label{sec:exp:rm-evaluation}
% \textbf{Calibration Metric of RM}
To study the uncertainty quantification ability of the reward model, we study 
ECE~\cite{ECE}, which is a metric used to assess model miscalibration. 
It involves binning assigned probability scores and comparing them to the average accuracies within these bins.
Following the Bradley–Terry model, the probability score of preferring an answer $\by^w$ over $\by^l$ can be calculated as:
\begin{equation}
\begin{aligned}
P(\by^w > \by^l| \bx) &= \frac{exp(r(\by^w | \bx))}{exp(r(\by^w | \bx))+ exp(r(\by^l | \bx))} \\
 &= \frac{1}{1+ exp(r(\by^w | \bx)-r(\by^l | \bx))}
\label{equation:rm-probability}
\end{aligned}
\end{equation}
Then we can define the Expected Calibration Error (ECE) for the reward model:
\begin{equation}
\begin{aligned}
\text{ECE} = \sum_m^M \frac{|B_m|}{\sum_m {|B_m|}} |\text{ACC}(B_m) - \text{CONF}(B_m)|,
\label{equation:rm-ece}
\end{aligned}
\end{equation}
where we divide samples into $M=15$ bins, $B_m$, according to the reward difference, and 
\begin{equation}
\begin{aligned}
\text{ACC}(B_m) = |B_m|^{-1} \sum_{i \in B_m} \mathbb{I} [r(\by_i^w | \bx) > r(\by_i^l | \bx)],  \\
\text{CONF}(B_m) = |B_m|^{-1} \sum_{i \in B_m} P(\by_i^w > \by_i^l| \bx),
\end{aligned}
\end{equation}
where $\mathbb{I}$ is the indicator function.
We observe that different reward models have different reward scales. 
To calculate ECE, we scale reward differences to ensure that the largest reward difference in the test dataset corresponds to $0.99$ confidence, which induces the calibrated ACC.
% More details for ECE of reward models can be found in Appendix~\ref{appendix:ece}.

\begin{table}[h]%
	\centering
 \caption{Accuracy and ECE of different training methods for reward modeling on two datasets. The best-performing values are highlighted. All ensemble methods have 5 members.}
  \vskip 0.1in
 \resizebox{\linewidth}{!}{
		\begin{tabular}{lccc}
			\toprule
			{Base Model}& {Training Method} & ACC $\uparrow$ & ECE $\downarrow$ \\
			\cmidrule(lr){1-4} % 调整中间横线长度
			\multirow{3}{*}{OPT-330M}&Full FT& 0.694 & 0.485 \\
            % &Full FT with MC Dropout & 0.689  &  \\
            % &Full FT Ensemble &   &  \\
			&LoRA Ensemble& \textbf{0.697} & \textbf{0.480} \\
			&Diverse LoRA Ensemble & \textbf{0.697}  & 0.481 \\
            % NNM=0.1, weight decay=0.1
			\cmidrule(lr){1-4} % 调整中间横线长度
			\multirow{3}{*}{Llama2-7B}& Full FT& 0.685 & 0.515  \\
            % &Full FT with MC Dropout &   & - \\
            % &Full FT Ensemble &   &  \\
			&LoRA Ensemble& 0.710 & 0.496 \\
			&Diverse LoRA Ensemble & \textbf{0.720} & \textbf{0.485} \\
			\bottomrule
		\end{tabular}
	}
	\label{tab:reward_model}

\end{table}

We establish reward models using OPT-330M on TL;DR and using Llama2-7B on the Anthropic Helpful dataset.
% We also evaluate different training methods.
Table~\ref{tab:reward_model} details the performance of reward models with different training methods and it can be observed that LoRA Ensemble benefits both accuracy and ECE on the test dataset.
Utilizing NNM, the overall performance in terms of the two metrics can be further improved.

\begin{figure}[h]
    \subfigure[]
    {
        \centering
        \includegraphics[width=0.45\linewidth]{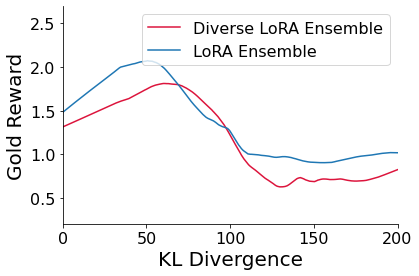}
    }
    \quad
    \subfigure[]
    {
        \centering
        \includegraphics[width=0.45\linewidth]{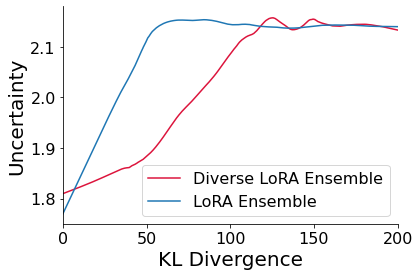}
    }
    \vspace{-1.5em}
    \caption{
    \label{fig:OOD-Detection}
    With diversity regularization, our proposed diverse reward LoRA ensemble achieves better OOD detection capabilities.
    \label{fig-step3-summarization}}
\end{figure}

We use two reward models, which are trained with LoRA ensemble and diverse LoRA ensemble to train the policy model respectively utilizing the RLHF objective ~\ref{equation:rlhf_reward}. 
Following ~\cite{gao2023scaling}, we utilize the KL divergence between the policy model and the SFT model $\mathrm{D_{KL}} \left( \pi_\theta(\by |\bx) || \pi_\text{SFT}(\by |\bx) \right)$ to measure the degree of policy optimization.
As shown in Figure~\ref{fig:OOD-Detection}, the uncertainty provided by the reward LoRA ensemble grows rapidly in the range of KL divergence from 0 to 50, which makes it difficult to distinguish between samples with high gold rewards and samples generated by over-optimized models (KL divergence roughly from 50 to 100).
On the contrary, our proposed diverse reward LoRA ensemble provides gradually increased uncertainty as the optimization process, indicating better OOD detection capabilities.

% 解释一下本来就很不自信了，所以ECE差不多

% \subsection{KL Regularization and Uncertainty Regularization}

% \subsubsection{Results}

% \subsection{Ablation Studies}
% In this section, we perform extra ablations to gain a better understanding of the effect of each component on the performance.

% \textbf{Effect of NNM Weight $\lambda$.}
% In Section~\ref{sec:exp:rm-evaluation}, we observe that Diverse LoRA Ensemblecan improve the uncertainty quantification ability of reward models.
% In this section, we first study the effect of NNM weights on reward modeling.

% Moving to RL fine-tuning, different RMs induce different policy models.

% \begin{table}[h]
% 	\centering
% 	%\resizebox{.95\columnwidth}{!}{
% 		\resizebox{0.48\textwidth}{!}{
% 			\begin{tabular}{l  c c  c c c c c c}
% 				\toprule
% 				&-10&-1 &-0.1&-0.01&0.01&0.1&1&10   \\ 
% 				\midrule
% 				H-R &27.7& 39.1 & 40.6 &38.5&40.2& 41.2&41.6&26.0  \\
% 				W-M-R &43.7 & 85.5  & 91.1&91.3&91.8& 88.6& 8.6&0.1 \\ 
% 				\bottomrule
% 				%		0.05 & 0.83 & 0.75 &  & &\\ \hline
% 			\end{tabular}
% 		}
% 		\caption{
% 			Ablation of the different optimism hyper-parameter $\lambda^o$ for ORPO.
% 			``H-R'' represents HalfCheetah-Random-v2 and ``W-M-R'' represents Walker2d-Medium-Replay-v2.}
% 		\label{table:lambda}
% 		%		\vspace{-1em}
% 	\end{table}

\subsection{Effect of Uncertainty Penalty}
\begin{figure}[h]
    \subfigure[ Dashed lines represent RM scores and solid lines represent gold RM scores.
    % by reward models and gold reward models.
        \label{fig-UP-overoptimization}]
        {
            \centering
            \includegraphics[width=0.45\linewidth]{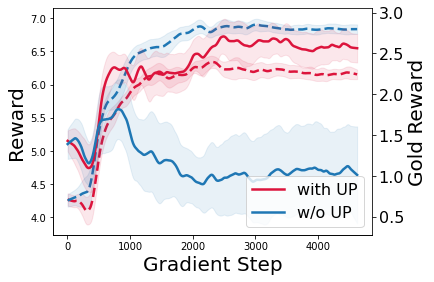}
        }
        \quad
    \subfigure[Reward uncertainty of LoRA ensembles.
    \label{fig-UP-uncertainty}]
    {
        \centering
        \includegraphics[width=0.45\linewidth]{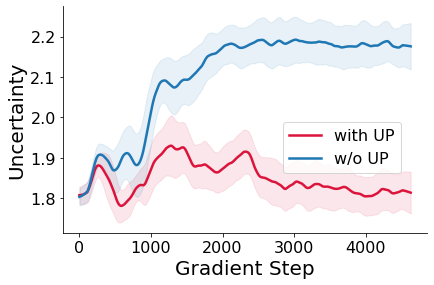}
    }
    \vspace{-1em}
    \caption{
    Uncertainty penalty ablation on policy model evaluation in the summarization task over 4 different seeds.
    \label{fig-step3-summarization}}
\end{figure}

Even with diverse reward LoRA ensembles, we observe significant overoptimization during the optimization process to the mean reward of the ensembles, as shown in Figure~\ref{fig-UP-overoptimization}.
When incorporating uncertainty penalties into rewards, the uncertainty of generated samples is well-controlled within a reasonable range, and the overoptimization issue is eliminated. 
This demonstrates the effectiveness of uncertainty regularization in mitigating overoptimization.

Interestingly, we observe that though utilizing uncertainty regularization can improve the overall performance in terms of gold RM, the RM score is diminished. 
This may be because uncertainty-penalized rewards limit the exploration of OOD output by the policy model, whether these outputs are high-quality or low-quality.
% Besides, for high-quality samples, the KL regularization is also strong. 
In this case, using additional uncertainty regularization may restrict the exploration of policy models, which corresponds to the exploration-exploitation dilemma in RL.

% TODO: 改成四幅图： uncertainty, KL, reward, Gold reward

% \textbf{Effect of KL Regularization.}
% To overcome the shortcomings of the KL penalty in RLHF and avoid mutual interference between KL and uncertainty penalties, we make the KL regularization independent of the actor loss in UP-RLHF.
% We compare the performance of using KL regularization with KL penalties employed by RLHF.

\subsection{Policy Model Evaluation}
% \subsubsection{TL;DR}
In this section, we compare our proposed UP-RLHF with existing RLHF methods in both summarization and question-answering tasks.
We compare gold RM scores instead of RM scores because different RMs have different scaling, thus making no sense to compare RM scores directly.

\begin{figure}[h]
    \subfigure[OPT-1.3B in the summarization task over 4 seeds.
        \label{fig-ORPO}]
    {
        \centering
        \includegraphics[width=0.45\linewidth]{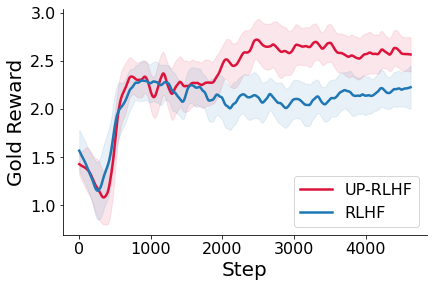}
    }
    \quad
    \subfigure[OPT-1.3B in the summarization task over 4 seeds.
    \label{fig-P-MDP}]
    {
        \centering
        \includegraphics[width=0.45\linewidth]{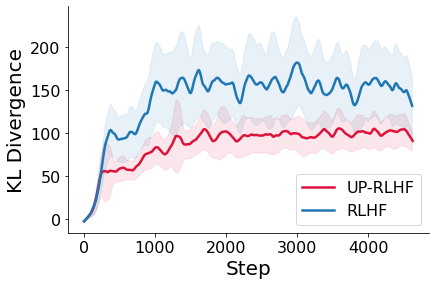}
    }
    \subfigure[Llama2-7B in the question-answering task.
    \label{fig-ORPO}]
    {
        \centering
        \includegraphics[width=0.45\linewidth]{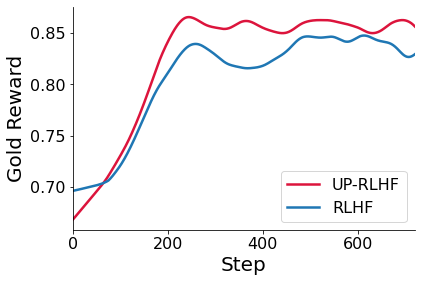}
    }
    \quad
    \subfigure[Llama2-7B in the question-answering task.
    \label{fig-P-MDP}]
    {
        \centering
        \includegraphics[width=0.45\linewidth]{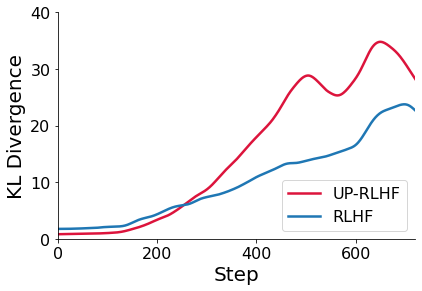}
    }
    \vspace{-1em}
    \caption{
    Comparison of UP-RLHF and RLHF.
    \label{fig-step3-performance}}
\end{figure}

As shown in Figure~\ref{fig-step3-performance}, UP-RLHF outperforms RLHF in terms of gold performance with a large margin in both tasks.
Especially in the summarization task, compared to RLHF, UP-RLHF can achieve higher performance with less KL divergence cost.
Note that the RLHF method utilized the full fine-tuning for reward modeling, while our diverse reward LoRA ensemble in UP-RLHF only fine-tunes 4.53\% parameters for OPT-350M and 1.25\% parameters for Llama2-7B.  

% \subsubsection{HH}
% Besides, deriving from objective~\ref{equation:rlhf_reward}, DPO~\cite{DPO} leverages rank loss to bypass the need for RL and explicit reward modeling.

% \subsubsection{GPT-4 Evaluation}

% \subsubsection{Human Evaluation} 
% For this evaluation, we

\section{Related Works}
\subsection{Reinforcement Learning from Human Feedback}
%RLHF ->  替代方法 offlineRL、 rank loss、 reject sampling finetuning
% RL和非RL，非RL又分online和offline
% overoptimization
RLHF is a pivotal approach for fine-tuning language models to align with human preferences. 
% In this context, the agent learns human preference from ranked answers provided by human annotators. 
% RLHF leverages RL to achieve this goal, relying on reward modeling as human proxies. 
Researchers have applied RLHF to diverse tasks~\cite{NLPO} such as text summarization~\cite{stiennon2020learning} and enhancing the harmlessness and helpfulness of language models~\cite{bai2022training}. Notably, InstructGPT introduces the three-step RLHF pipeline using a supervised approach and the PPO algorithm~\cite{PPO}, demonstrating its effectiveness on ChatGPT.
While successful, RLHF faces various challenges~\cite{casper2023open}.
One of the most pressing challenges is overoptimization, which is caused by imperfect RMs~\cite{gao2023scaling}.
The author in~\cite{gao2023scaling} provides the scaling law of RMs, which shows the effect of increasing RM parameters and data size in mitigating the issue.

RLHF heavily relies on reward modeling to proxy human preferences.
Some recent works aim to bypass the reward modeling step~\cite{RRHF,DPO,PRO}.
Specifically, DPO directly optimizes the policy towards the objective~\ref{equation:rlhf_reward} by solving a classification problem on the human preference data.
Although bypassing the reward modeling step benefits from easy implementation and training stability, more recent works reveal several advantages of using reward models.
\cite{IPO} analyzes the robustness of reward-model-based methods against overfitting caused by the weakness of the KL regularization.
% From the RL review, \cite{sun2023reinforcement} advocates that as long as inverse RL is better than imitation learning, RLHF is better than SFT.
Besides, compared to DPO, reward-model-based RLHF shows great advantages on out-of-preference samples~\cite{ReMax, li2023policy}.

There are many works to address the challenge in RLHF such as computational overhead~\cite{ReMax}, sample efficiency~\cite{ILQL,ReST}, unstable training~\cite{P3O}, and overoptimization~\cite{moskovitz2023confronting,coste2023reward, eisenstein2023helping}.
We also focus on the overoptimization issue.
While most recent works focus only on the RL fine-tuning step, we first introduce uncertainty quantification to the reward modeling step and make the RL fine-tuning uncertainty aware.
% There are also many works to address the challenge in RLHF such as computational overhead, sample efficiency~\cite{choshen2019weaknesses}, and unstable training~\cite{casper2023open, Moss}.
% Numerous studies have explored alternatives for RLHF.
% ReMax~\cite{ReMax} replaces PPO in RLHF with REINFORCE~\cite{REINFORCE}, simplifying the implementation and reducing memory usage by 50\%.
% ILQL~\cite{ILQL} and ReST~\cite{ReST} utilize offline RL algorithms to improve the sample efficiency.
% To improve the training stability, P3O~\cite{P3O} introduces a trajectory-wise policy gradient algorithm based on PPO, which is invariant to equivalent reward functions trained with human preferences.

% Recent works also aim to learn from feedback with non-RL approaches.
% For instance, Quark~\cite{Quark}, RAFT~\cite{RAFT}, and Llama2~\cite{Llama2} filter samples with high reward by reject sampling to fine-tune the model.
% RRHF~\cite{RRHF}, DPO~\cite{DPO}, and PRO~\cite{PRO} utilize the rank loss to achieve the RLHF objective.

\subsection{Uncertainty Aware Reinforcement Learning}
Uncertainty is a pivotal factor in the realm of RL. 
The Optimism in the Face of Uncertainty (OFU) principle~\cite{bandit-2011} in online RL strategies is widely adopted for facilitating active and efficient exploration of the environment~\cite{lockwood2022review}. 
In offline RL~\cite{levine2020offline}, uncertainty is typically utilized for conservative to control the prediction errors caused by imperfect dynamics models.
Uncertainty is usually estimated by value networks in model-free RL~\cite{Disagreemet, PBRL} and by dynamics models in model-based RL~\cite{MBPO, MOPO}.

RLHF can be formulated as reverse RL with offline datasets, where reward models trained on an offline limited preference dataset are imperfect. 
Inspired by recent model-based offline RL methods~\cite{MOPO, MOReL, lu2022revisiting}, we propose to penalize rewards with the model uncertainty for conservative policy optimization, aiming for mitigating the overoptimization issue.
Concurrent work by~\cite{coste2023reward, eisenstein2023helping} also shows reward model ensemble helps mitigate overoptimization.
However, utilizing reward model ensembles increases RM parameters several times, and may lack diversity between ensemble members~\cite{gleave2022uncertainty}.
To diversify reward ensembles, ~\cite{eisenstein2023helping} propose to use different seeds in the pre-training phase.
We propose to train diverse LoRA ensembles with NNM for reward modeling, which is much cheaper and parameter-effective.
Besides, we analyze the relations between KL and uncertainty regularization and make them affect independently.
% , and evaluate our method on real human preference datasets.

\subsection{Uncertainty for LLMs}
Uncertainty quantification for deep neural networks has been well studied~\cite{gawlikowski2023survey}.
Popular methods include deep ensemble, MC dropout~\cite{gal2016dropout}, and so on.
In the context of LLMs, some new challenges arise.
Diversity plays an important role in ensemble-based methods~\cite{breiman2001random}.
However, fine-tuning LLMs for ensembles~\cite{sun2022quantifying} not only is too expensive to scale up but also lacks diversity~\cite{gleave2022uncertainty}.
Therefore, we adopt a popular PEFT technology, LoRA~\cite{LoRA} for training the ensemble of reward models.
Different from the concurrent work~\cite{wang2023loraEnsemble} which also proposes LoRA ensemble for LLMs fine-tuning and different regularization techniques for each LoRA, we propose a diversity regularization to encourage diversity between ensemble members.
Besides, we mainly focus on the reward modeling in the context of RLHF.

% \subsection{Uncertainty regularization and KL regularization.}
%对差的OOD样本的处理能力，KL约束很弱，但RM能展示出强大的不确定性

\section{Conclusion and Limitations}
In this paper, we propose UP-RLHF, an uncertainty-aware RLHF framework that contributes to the uncertainty of AI systems based on LLMs.
Our proposed diverse reward LoRA ensemble can provide satisfactory uncertainty quantification for samples in RLHF.
Leveraging the reward uncertainty, we highlight the pivotal role of uncertainty regularization in effectively addressing the overoptimization challenge in the alignment of LLMs.

Our work has limitations. 
While the diverse reward LoRA ensemble proves to be parameter-effective, the computation of the nuclear norm for concatenated LoRA matrices introduces additional time overhead. Moreover, uncertainty regularization may exhibit over-conservatism, particularly in cases involving near-distribution high-quality outputs. As a future direction, exploring methods to strike a balance between KL and uncertainty regularization for specific samples could further refine the framework's performance.
%增加了Uncertainty regularization，也增大了对正样本的约束
% LoRA Ensemble 训reward model还是很贵，SVD也很慢

\nocite{langley00}

\bibliography{paper}

\begin{thebibliography}{54}
\providecommand{\natexlab}[1]{#1}
\providecommand{\url}[1]{\texttt{#1}}
\expandafter\ifx\csname urlstyle\endcsname\relax
  \providecommand{\doi}[1]{doi: #1}\else
  \providecommand{\doi}{doi: \begingroup \urlstyle{rm}\Url}\fi

\bibitem[Abbasi-Yadkori et~al.(2011)Abbasi-Yadkori, P{\'a}l, and
  Szepesv{\'a}ri]{bandit-2011}
Abbasi-Yadkori, Y., P{\'a}l, D., and Szepesv{\'a}ri, C.
\newblock Improved algorithms for linear stochastic bandits.
\newblock \emph{Advances in neural information processing systems}, 24, 2011.

\bibitem[Azar et~al.(2023)Azar, Rowland, Piot, Guo, Calandriello, Valko, and
  Munos]{IPO}
Azar, M.~G., Rowland, M., Piot, B., Guo, D., Calandriello, D., Valko, M., and
  Munos, R.
\newblock A general theoretical paradigm to understand learning from human
  preferences.
\newblock \emph{arXiv preprint arXiv:2310.12036}, 2023.

\bibitem[Bai et~al.(2022{\natexlab{a}})Bai, Wang, Yang, Deng, Garg, Liu, and
  Wang]{PBRL}
Bai, C., Wang, L., Yang, Z., Deng, Z., Garg, A., Liu, P., and Wang, Z.
\newblock Pessimistic bootstrapping for uncertainty-driven offline
  reinforcement learning.
\newblock \emph{International Conference on Learning Representations},
  2022{\natexlab{a}}.

\bibitem[Bai et~al.(2022{\natexlab{b}})Bai, Jones, Ndousse, Askell, Chen,
  DasSarma, Drain, Fort, Ganguli, Henighan, et~al.]{bai2022training}
Bai, Y., Jones, A., Ndousse, K., Askell, A., Chen, A., DasSarma, N., Drain, D.,
  Fort, S., Ganguli, D., Henighan, T., et~al.
\newblock Training a helpful and harmless assistant with reinforcement learning
  from human feedback.
\newblock \emph{arXiv preprint arXiv:2204.05862}, 2022{\natexlab{b}}.

\bibitem[Beeching et~al.(2023)Beeching, Belkada, Rasul, Tunstall, von Werra,
  Rajani, and Lambert]{beeching2023stackllama}
Beeching, E., Belkada, Y., Rasul, K., Tunstall, L., von Werra, L., Rajani, N.,
  and Lambert, N.
\newblock Stackllama: An rl fine-tuned llama model for stack exchange question
  and answering, 2023.
\newblock \emph{URL https://huggingface. co/blog/stackllama}, 1, 2023.

\bibitem[Bradley \& Terry(1952)Bradley and Terry]{bradley1952rank}
Bradley, R.~A. and Terry, M.~E.
\newblock Rank analysis of incomplete block designs: I. the method of paired
  comparisons.
\newblock \emph{Biometrika}, 39\penalty0 (3/4):\penalty0 324--345, 1952.

\bibitem[Breiman(2001)]{breiman2001random}
Breiman, L.
\newblock Random forests.
\newblock \emph{Machine learning}, 45:\penalty0 5--32, 2001.

\bibitem[Brown et~al.(2020)Brown, Mann, Ryder, Subbiah, Kaplan, Dhariwal,
  Neelakantan, Shyam, Sastry, Askell, et~al.]{2020GPT3}
Brown, T., Mann, B., Ryder, N., Subbiah, M., Kaplan, J.~D., Dhariwal, P.,
  Neelakantan, A., Shyam, P., Sastry, G., Askell, A., et~al.
\newblock Language models are few-shot learners.
\newblock \emph{Advances in neural information processing systems},
  33:\penalty0 1877--1901, 2020.

\bibitem[Casper et~al.(2023)Casper, Davies, Shi, Gilbert, Scheurer, Rando,
  Freedman, Korbak, Lindner, Freire, et~al.]{casper2023open}
Casper, S., Davies, X., Shi, C., Gilbert, T.~K., Scheurer, J., Rando, J.,
  Freedman, R., Korbak, T., Lindner, D., Freire, P., et~al.
\newblock Open problems and fundamental limitations of reinforcement learning
  from human feedback.
\newblock \emph{arXiv preprint arXiv:2307.15217}, 2023.

\bibitem[Coste et~al.(2023)Coste, Anwar, Kirk, and Krueger]{coste2023reward}
Coste, T., Anwar, U., Kirk, R., and Krueger, D.
\newblock Reward model ensembles help mitigate overoptimization.
\newblock \emph{arXiv preprint arXiv:2310.02743}, 2023.

\bibitem[Eisenstein et~al.(2023)Eisenstein, Nagpal, Agarwal, Beirami, D'Amour,
  Dvijotham, Fisch, Heller, Pfohl, Ramachandran, et~al.]{eisenstein2023helping}
Eisenstein, J., Nagpal, C., Agarwal, A., Beirami, A., D'Amour, A., Dvijotham,
  D., Fisch, A., Heller, K., Pfohl, S., Ramachandran, D., et~al.
\newblock Helping or herding? reward model ensembles mitigate but do not
  eliminate reward hacking.
\newblock \emph{arXiv preprint arXiv:2312.09244}, 2023.

\bibitem[Ethayarajh et~al.(2022)Ethayarajh, Choi, and
  Swayamdipta]{pmlr-v162-ethayarajh22a}
Ethayarajh, K., Choi, Y., and Swayamdipta, S.
\newblock Understanding dataset difficulty with $\mathcal{V}$-usable
  information.
\newblock In Chaudhuri, K., Jegelka, S., Song, L., Szepesvari, C., Niu, G., and
  Sabato, S. (eds.), \emph{Proceedings of the 39th International Conference on
  Machine Learning}, volume 162 of \emph{Proceedings of Machine Learning
  Research}, pp.\  5988--6008. PMLR, 17--23 Jul 2022.
\newblock URL \url{https://proceedings.mlr.press/v162/ethayarajh22a.html}.

\bibitem[Gal \& Ghahramani(2016)Gal and Ghahramani]{gal2016dropout}
Gal, Y. and Ghahramani, Z.
\newblock Dropout as a bayesian approximation: Representing model uncertainty
  in deep learning.
\newblock In \emph{international conference on machine learning}, pp.\
  1050--1059. PMLR, 2016.

\bibitem[Gao et~al.(2023)Gao, Schulman, and Hilton]{gao2023scaling}
Gao, L., Schulman, J., and Hilton, J.
\newblock Scaling laws for reward model overoptimization.
\newblock In \emph{International Conference on Machine Learning}, pp.\
  10835--10866. PMLR, 2023.

\bibitem[Gawlikowski et~al.(2023)Gawlikowski, Tassi, Ali, Lee, Humt, Feng,
  Kruspe, Triebel, Jung, Roscher, et~al.]{gawlikowski2023survey}
Gawlikowski, J., Tassi, C. R.~N., Ali, M., Lee, J., Humt, M., Feng, J., Kruspe,
  A., Triebel, R., Jung, P., Roscher, R., et~al.
\newblock A survey of uncertainty in deep neural networks.
\newblock \emph{Artificial Intelligence Review}, 56\penalty0 (Suppl
  1):\penalty0 1513--1589, 2023.

\bibitem[Gleave \& Irving(2022)Gleave and Irving]{gleave2022uncertainty}
Gleave, A. and Irving, G.
\newblock Uncertainty estimation for language reward models.
\newblock \emph{arXiv preprint arXiv:2203.07472}, 2022.

\bibitem[Go et~al.(2023)Go, Korbak, Kruszewski, Rozen, Ryu, and
  Dymetman]{go2023aligning}
Go, D., Korbak, T., Kruszewski, G., Rozen, J., Ryu, N., and Dymetman, M.
\newblock Aligning language models with preferences through f-divergence
  minimization.
\newblock In \emph{International conference on machine learning}. PMLR, 2023.

\bibitem[Gulcehre et~al.(2023)Gulcehre, Paine, Srinivasan, Konyushkova, Weerts,
  Sharma, Siddhant, Ahern, Wang, Gu, et~al.]{ReST}
Gulcehre, C., Paine, T.~L., Srinivasan, S., Konyushkova, K., Weerts, L.,
  Sharma, A., Siddhant, A., Ahern, A., Wang, M., Gu, C., et~al.
\newblock Reinforced self-training (rest) for language modeling.
\newblock \emph{arXiv preprint arXiv:2308.08998}, 2023.

\bibitem[Hu et~al.(2022)Hu, Wallis, Allen-Zhu, Li, Wang, Wang, Chen,
  et~al.]{LoRA}
Hu, E.~J., Wallis, P., Allen-Zhu, Z., Li, Y., Wang, S., Wang, L., Chen, W.,
  et~al.
\newblock Lora: Low-rank adaptation of large language models.
\newblock In \emph{International Conference on Learning Representations}, 2022.

\bibitem[Janner et~al.(2019)Janner, Fu, Zhang, and Levine]{MBPO}
Janner, M., Fu, J., Zhang, M., and Levine, S.
\newblock When to trust your model: Model-based policy optimization.
\newblock \emph{Advances in Neural Information Processing Systems}, 32, 2019.

\bibitem[Kidambi et~al.(2020)Kidambi, Rajeswaran, Netrapalli, and
  Joachims]{MOReL}
Kidambi, R., Rajeswaran, A., Netrapalli, P., and Joachims, T.
\newblock Morel: Model-based offline reinforcement learning.
\newblock \emph{Advances in neural information processing systems},
  33:\penalty0 21810--21823, 2020.

\bibitem[Korbak et~al.(2022)Korbak, Elsahar, Kruszewski, and
  Dymetman]{korbak2022reinforcement}
Korbak, T., Elsahar, H., Kruszewski, G., and Dymetman, M.
\newblock On reinforcement learning and distribution matching for fine-tuning
  language models with no catastrophic forgetting.
\newblock \emph{Advances in Neural Information Processing Systems},
  35:\penalty0 16203--16220, 2022.

\bibitem[Kreps et~al.(2022)Kreps, McCain, and Brundage]{kreps2022all}
Kreps, S., McCain, R.~M., and Brundage, M.
\newblock All the news that’s fit to fabricate: Ai-generated text as a tool
  of media misinformation.
\newblock \emph{Journal of experimental political science}, 9\penalty0
  (1):\penalty0 104--117, 2022.

\bibitem[Levine et~al.(2020)Levine, Kumar, Tucker, and Fu]{levine2020offline}
Levine, S., Kumar, A., Tucker, G., and Fu, J.
\newblock Offline reinforcement learning: Tutorial, review, and perspectives on
  open problems.
\newblock \emph{arXiv preprint arXiv:2005.01643}, 2020.

\bibitem[Li et~al.(2023{\natexlab{a}})Li, Xu, and Yu]{li2023policy}
Li, Z., Xu, T., and Yu, Y.
\newblock Policy optimization in rlhf: The impact of out-of-preference data.
\newblock \emph{arXiv preprint arXiv:2312.10584}, 2023{\natexlab{a}}.

\bibitem[Li et~al.(2023{\natexlab{b}})Li, Xu, Zhang, Yu, Sun, and Luo]{ReMax}
Li, Z., Xu, T., Zhang, Y., Yu, Y., Sun, R., and Luo, Z.
\newblock Remax: A simple, effective, and efficient method for aligning large
  language models.
\newblock \emph{arXiv preprint arXiv:2310.10505}, 2023{\natexlab{b}}.

\bibitem[Lockwood \& Si(2022)Lockwood and Si]{lockwood2022review}
Lockwood, O. and Si, M.
\newblock A review of uncertainty for deep reinforcement learning.
\newblock In \emph{Proceedings of the AAAI Conference on Artificial
  Intelligence and Interactive Digital Entertainment}, volume~18, pp.\
  155--162, 2022.

\bibitem[Lu et~al.(2022)Lu, Ball, Parker-Holder, Osborne, and
  Roberts]{lu2022revisiting}
Lu, C., Ball, P., Parker-Holder, J., Osborne, M., and Roberts, S.~J.
\newblock Revisiting design choices in offline model based reinforcement
  learning.
\newblock In \emph{International Conference on Learning Representations}, 2022.

\bibitem[Mnih et~al.(2016)Mnih, Badia, Mirza, Graves, Lillicrap, Harley,
  Silver, and Kavukcuoglu]{A2C}
Mnih, V., Badia, A.~P., Mirza, M., Graves, A., Lillicrap, T., Harley, T.,
  Silver, D., and Kavukcuoglu, K.
\newblock Asynchronous methods for deep reinforcement learning.
\newblock In \emph{International conference on machine learning}, pp.\
  1928--1937. PMLR, 2016.

\bibitem[Moskovitz et~al.(2023)Moskovitz, Singh, Strouse, Sandholm,
  Salakhutdinov, Dragan, and McAleer]{moskovitz2023confronting}
Moskovitz, T., Singh, A.~K., Strouse, D., Sandholm, T., Salakhutdinov, R.,
  Dragan, A.~D., and McAleer, S.
\newblock Confronting reward model overoptimization with constrained rlhf.
\newblock \emph{arXiv preprint arXiv:2310.04373}, 2023.

\bibitem[Naeini et~al.(2015)Naeini, Cooper, and Hauskrecht]{ECE}
Naeini, M.~P., Cooper, G., and Hauskrecht, M.
\newblock Obtaining well calibrated probabilities using bayesian binning.
\newblock In \emph{Proceedings of the AAAI conference on artificial
  intelligence}, volume~29, 2015.

\bibitem[Ouyang et~al.(2022)Ouyang, Wu, Jiang, Almeida, Wainwright, Mishkin,
  Zhang, Agarwal, Slama, Ray, et~al.]{InstructGPT}
Ouyang, L., Wu, J., Jiang, X., Almeida, D., Wainwright, C., Mishkin, P., Zhang,
  C., Agarwal, S., Slama, K., Ray, A., et~al.
\newblock Training language models to follow instructions with human feedback.
\newblock \emph{Advances in Neural Information Processing Systems},
  35:\penalty0 27730--27744, 2022.

\bibitem[Pathak et~al.(2019)Pathak, Gandhi, and Gupta]{Disagreemet}
Pathak, D., Gandhi, D., and Gupta, A.
\newblock Self-supervised exploration via disagreement.
\newblock In \emph{International Conference on Machine Learning}, pp.\
  5062--5071. PMLR, 2019.

\bibitem[Peng et~al.(2019)Peng, Kumar, Zhang, and Levine]{AWR}
Peng, X., Kumar, A., Zhang, G., and Levine, S.
\newblock Advantage-weighted regression: Simple and scalable off-policy
  reinforcement learning.
\newblock \emph{arXiv preprint arXiv:1910.00177}, 2019.

\bibitem[Perez et~al.(2022)Perez, Huang, Song, Cai, Ring, Aslanides, Glaese,
  McAleese, and Irving]{perez2022red}
Perez, E., Huang, S., Song, F., Cai, T., Ring, R., Aslanides, J., Glaese, A.,
  McAleese, N., and Irving, G.
\newblock Red teaming language models with language models.
\newblock In \emph{Proceedings of the 2022 Conference on Empirical Methods in
  Natural Language Processing}, pp.\  3419--3448, 2022.

\bibitem[Rafailov et~al.(2023)Rafailov, Sharma, Mitchell, Ermon, Manning, and
  Finn]{DPO}
Rafailov, R., Sharma, A., Mitchell, E., Ermon, S., Manning, C.~D., and Finn, C.
\newblock Direct preference optimization: Your language model is secretly a
  reward model.
\newblock \emph{Advances in Neural Information Processing Systems}, 2023.

\bibitem[Ramamurthy et~al.(2023)Ramamurthy, Ammanabrolu, Brantley, Hessel,
  Sifa, Bauckhage, Hajishirzi, and Choi]{NLPO}
Ramamurthy, R., Ammanabrolu, P., Brantley, K., Hessel, J., Sifa, R., Bauckhage,
  C., Hajishirzi, H., and Choi, Y.
\newblock Is reinforcement learning (not) for natural language processing:
  Benchmarks, baselines, and building blocks for natural language policy
  optimization.
\newblock In \emph{The Eleventh International Conference on Learning
  Representations}, 2023.

\bibitem[Schulman et~al.(2015)Schulman, Levine, Abbeel, Jordan, and
  Moritz]{TRPO}
Schulman, J., Levine, S., Abbeel, P., Jordan, M., and Moritz, P.
\newblock Trust region policy optimization.
\newblock In \emph{International conference on machine learning}, pp.\
  1889--1897. PMLR, 2015.

\bibitem[Schulman et~al.(2017)Schulman, Wolski, Dhariwal, Radford, and
  Klimov]{PPO}
Schulman, J., Wolski, F., Dhariwal, P., Radford, A., and Klimov, O.
\newblock Proximal policy optimization algorithms.
\newblock \emph{arXiv preprint arXiv:1707.06347}, 2017.

\bibitem[Snell et~al.(2023)Snell, Kostrikov, Su, Yang, and Levine]{ILQL}
Snell, C.~V., Kostrikov, I., Su, Y., Yang, S., and Levine, S.
\newblock Offline rl for natural language generation with implicit language q
  learning.
\newblock In \emph{The Eleventh International Conference on Learning
  Representations}, 2023.

\bibitem[Song et~al.(2023)Song, Yu, Li, Yu, Huang, Li, and Wang]{PRO}
Song, F., Yu, B., Li, M., Yu, H., Huang, F., Li, Y., and Wang, H.
\newblock Preference ranking optimization for human alignment.
\newblock \emph{arXiv preprint arXiv:2306.17492}, 2023.

\bibitem[Stiennon et~al.(2020)Stiennon, Ouyang, Wu, Ziegler, Lowe, Voss,
  Radford, Amodei, and Christiano]{stiennon2020learning}
Stiennon, N., Ouyang, L., Wu, J., Ziegler, D., Lowe, R., Voss, C., Radford, A.,
  Amodei, D., and Christiano, P.~F.
\newblock Learning to summarize with human feedback.
\newblock \emph{Advances in Neural Information Processing Systems},
  33:\penalty0 3008--3021, 2020.

\bibitem[Sun et~al.(2022)Sun, Yan, Abbeel, and Mordatch]{sun2022quantifying}
Sun, M., Yan, W., Abbeel, P., and Mordatch, I.
\newblock Quantifying uncertainty in foundation models via ensembles.
\newblock In \emph{NeurIPS 2022 Workshop on Robustness in Sequence Modeling},
  2022.

\bibitem[Touvron et~al.(2023)Touvron, Martin, Stone, Albert, Almahairi, Babaei,
  Bashlykov, Batra, Bhargava, Bhosale, et~al.]{Llama2}
Touvron, H., Martin, L., Stone, K., Albert, P., Almahairi, A., Babaei, Y.,
  Bashlykov, N., Batra, S., Bhargava, P., Bhosale, S., et~al.
\newblock Llama 2: Open foundation and fine-tuned chat models.
\newblock \emph{arXiv preprint arXiv:2307.09288}, 2023.

\bibitem[Wang et~al.(2023)Wang, Aitchison, and Rudolph]{wang2023loraEnsemble}
Wang, X., Aitchison, L., and Rudolph, M.
\newblock Lora ensembles for large language model fine-tuning.
\newblock \emph{arXiv preprint arXiv:2310.00035}, 2023.

\bibitem[Williams(1992)]{REINFORCE}
Williams, R.~J.
\newblock Simple statistical gradient-following algorithms for connectionist
  reinforcement learning.
\newblock \emph{Machine learning}, 8:\penalty0 229--256, 1992.

\bibitem[Wu et~al.(2023)Wu, Zhu, Zhang, Wen, Ramchandran, and Jiao]{P3O}
Wu, T., Zhu, B., Zhang, R., Wen, Z., Ramchandran, K., and Jiao, J.
\newblock Pairwise proximal policy optimization: Harnessing relative feedback
  for llm alignment.
\newblock \emph{arXiv preprint arXiv:2310.00212}, 2023.

\bibitem[Yang et~al.(2023)Yang, Xiao, Wang, Zhang, Bian, Yin, Lv, Pan, Wang,
  Yan, et~al.]{Baichuan2}
Yang, A., Xiao, B., Wang, B., Zhang, B., Bian, C., Yin, C., Lv, C., Pan, D.,
  Wang, D., Yan, D., et~al.
\newblock Baichuan 2: Open large-scale language models.
\newblock \emph{arXiv preprint arXiv:2309.10305}, 2023.

\bibitem[Yao et~al.(2023)Yao, Aminabadi, Ruwase, Rajbhandari, Wu, Awan, Rasley,
  Zhang, Li, Holmes, et~al.]{yao2023deepspeed}
Yao, Z., Aminabadi, R.~Y., Ruwase, O., Rajbhandari, S., Wu, X., Awan, A.~A.,
  Rasley, J., Zhang, M., Li, C., Holmes, C., et~al.
\newblock Deepspeed-chat: Easy, fast and affordable rlhf training of
  chatgpt-like models at all scales.
\newblock \emph{arXiv preprint arXiv:2308.01320}, 2023.

\bibitem[Yu et~al.(2020)Yu, Thomas, Yu, Ermon, Zou, Levine, Finn, and Ma]{MOPO}
Yu, T., Thomas, G., Yu, L., Ermon, S., Zou, J.~Y., Levine, S., Finn, C., and
  Ma, T.
\newblock Mopo: Model-based offline policy optimization.
\newblock \emph{Advances in Neural Information Processing Systems},
  33:\penalty0 14129--14142, 2020.

\bibitem[Yuan et~al.(2023)Yuan, Yuan, Tan, Wang, Huang, and Huang]{RRHF}
Yuan, Z., Yuan, H., Tan, C., Wang, W., Huang, S., and Huang, F.
\newblock Rrhf: Rank responses to align language models with human feedback
  without tears.
\newblock \emph{Advances in neural information processing systems}, 2023.

\bibitem[Zhang et~al.(2022)Zhang, Roller, Goyal, Artetxe, Chen, Chen, Dewan,
  Diab, Li, Lin, et~al.]{OPT}
Zhang, S., Roller, S., Goyal, N., Artetxe, M., Chen, M., Chen, S., Dewan, C.,
  Diab, M., Li, X., Lin, X.~V., et~al.
\newblock Opt: Open pre-trained transformer language models.
\newblock \emph{arXiv preprint arXiv:2205.01068}, 2022.

\bibitem[Zhu et~al.(2023)Zhu, Sharma, Frujeri, Dong, Zhu, Jordan, and
  Jiao]{zhu2023fine}
Zhu, B., Sharma, H., Frujeri, F.~V., Dong, S., Zhu, C., Jordan, M.~I., and
  Jiao, J.
\newblock Fine-tuning language models with advantage-induced policy alignment.
\newblock \emph{arXiv preprint arXiv:2306.02231}, 2023.

\bibitem[Ziegler et~al.(2019)Ziegler, Stiennon, Wu, Brown, Radford, Amodei,
  Christiano, and Irving]{2019TLDR}
Ziegler, D.~M., Stiennon, N., Wu, J., Brown, T.~B., Radford, A., Amodei, D.,
  Christiano, P., and Irving, G.
\newblock Fine-tuning language models from human preferences.
\newblock \emph{arXiv preprint arXiv:1909.08593}, 2019.

\end{thebibliography}
\bibliographystyle{icml2024}

%%%%%%%%%%%%%%%%%%%%%%%%%%%%%%%%%%%%%%%%%%%%%%%%%%%%%%%%%%%%%%%%%%%%%%%%%%%%%%%
%%%%%%%%%%%%%%%%%%%%%%%%%%%%%%%%%%%%%%%%%%%%%%%%%%%%%%%%%%%%%%%%%%%%%%%%%%%%%%%
% APPENDIX
%%%%%%%%%%%%%%%%%%%%%%%%%%%%%%%%%%%%%%%%%%%%%%%%%%%%%%%%%%%%%%%%%%%%%%%%%%%%%%%
%%%%%%%%%%%%%%%%%%%%%%%%%%%%%%%%%%%%%%%%%%%%%%%%%%%%%%%%%%%%%%%%%%%%%%%%%%%%%%%
\newpage
\appendix
\onecolumn

\end{document}